\documentclass{article} % For LaTeX2e
\usepackage{CJKutf8} % 先加载 CJK，避免后续包改写断词宏产生提示
\usepackage{enumitem}
\usepackage{relook,times}

\usepackage{wrapfig}
\usepackage{lipsum} % 用于生成示例文字
\usepackage{wrapfig}

\usepackage{amsmath,amsfonts,bm}
\usepackage{multicol}
\def\eqref#1{equation~\ref{#1}}
\def\1{\bm{1}}

\DeclareMathAlphabet{\mathsfit}{\encodingdefault}{\sfdefault}{m}{sl}
\SetMathAlphabet{\mathsfit}{bold}{\encodingdefault}{\sfdefault}{bx}{n}

\usepackage{booktabs}      % 提供更美观的表格线 (如 \toprule, \midrule, \bottomrule)
\usepackage{multirow}      % 用于创建跨多行的单元格
\usepackage{amssymb}       % 提供了丰富的数学符号，包括 \checkmark (对勾)
\usepackage{graphicx}      % 用于处理图片和缩放内容 (\resizebox)
\usepackage{algorithm}     % 提供 algorithm 浮动环境
\usepackage{algorithmic}   % 提供算法伪代码排版命令（全大写命令）

\usepackage{colortbl}
\usepackage{xcolor}

\usepackage{hyperref}
\usepackage{url}
\usepackage{multicol}
\usepackage{aliascnt}
\usepackage{adjustbox}
\usepackage{tcolorbox}
\usepackage{siunitx} % for S column type in tables
\usepackage{booktabs}
\usepackage{multirow} 
\usepackage{array}
\usepackage{enumitem}
\usepackage{booktabs} % 用于更美观的表格线
\usepackage{amssymb}  % 用于 \checkmark
\usepackage{amsmath}  % 用于 \texttimes (或者使用 pifont 和 \ding{55})
\usepackage{graphicx} % 可能需要，如果调整列宽
\usepackage{longtable} % 如果表格太长需要跨页
\usepackage{placeins}
\usepackage{caption}
\usepackage{marvosym}
\definecolor{uclablue}{rgb}{0.15, 0.45, 0.68}
\hypersetup{
    breaklinks,
    citecolor=uclablue,
    colorlinks=true,
}
\title{ReLook: Vision-Grounded RL with a Multimodal LLM Critic for Agentic Web Coding}

\author{
Yuhang Li$^{1,*}$,
Chenchen Zhang$^{1,*,\dagger}$,
Ruilin Lv$^{2}$,
Ao Liu$^{1}$,
Ken Deng$^{2}$,
Yuanxing Zhang$^{3}$, \\
Jiaheng Liu$^{4}$,
Wiggin Zhou$^{1,\dagger}$,
Bo Zhou$^{1,\dagger}$\\[0.5em]
\textbf{$^1$LLM Department, Tencent} \quad \textbf{$^2$Independent Researcher}\\
\textbf{$^3$Peking University} \quad \textbf{$^4$Nanjing University}\\
}

\begin{document}
\maketitle
\let\oldthefootnote\thefootnote

% Equal contribution footnote
\let\thefootnote\relax\footnotetext{$^*$ The first two authors contributed equally to this work.}
% Corresponding authors footnote
\let\thefootnote\relax\footnotetext{$^\dagger$ Corresponding authors. \Letter~\{adamwzhang,wigginzhou,chaysezhou\}@tencent.com}
\let\thefootnote\oldthefootnote

\begin{abstract}
While Large Language Models (LLMs) excel at algorithmic code generation, they struggle with front-end development, where correctness is judged on rendered pixels and interaction. We present \textbf{ReLook}, an \emph{agentic}, vision-grounded reinforcement learning framework that empowers an \emph{agent} to close a robust generate--diagnose--refine loop by invoking a multimodal LLM (MLLM) as a tool.
During training, the agent uses the MLLM-in-the-loop both as a visual critic—scoring code with screenshots—and as a source of actionable, vision-grounded feedback; a strict zero-reward rule for invalid renders anchors renderability and prevents reward hacking. To prevent behavioral collapse, we introduce \emph{Forced Optimization}, a strict acceptance rule that admits only improving revisions, yielding monotonically better trajectories. At inference, we decouple the critic and run a lightweight, critic-free self-edit cycle, keeping latency \emph{comparable to base decoding} while retaining most of the gains.
Across three widely used benchmarks, ReLook consistently outperforms strong baselines in vision-grounded front-end code generation, highlighting the benefits of agentic perception, visual rewards, and training–inference decoupling.
\end{abstract}

\section{Introduction}

Large Language Models (LLMs) excel on closed-form benchmarks—programming contests~\citep{li2022competition}, SQL synthesis~\citep{liu2024survey}, and mathematical reasoning~\citep{yang2024qwen2, Deng2025HiPOHP}—yet still underperform on front-end code generation, where visual fidelity and interaction are first-class. Unlike binary unit tests in algorithmic tasks, front-end quality lies on a continuum: a single misaligned pixel can signify failure.

This perceptual barrier explains current shortcomings: text-only models are blind to pixel-level consequences, yielding (i) layout drift, (ii) interaction breakage, and (iii) aesthetic inconsistency. To address this, a model must (1) see rendered HTML/CSS/JS/SVG, (2) diagnose misalignments and broken interactions, and (3) iteratively refine in situ. Existing methods miss this loop: one-shot vision-to-code systems (pix2code~\citep{wust2024pix2code}, Design2Code~\citep{si2024design2code}, UICoder~\citep{wu2024uicoder}) generate but do not refine; self-refinement frameworks (CodeRL~\citep{le2022coderl}, Self-Refine~\citep{madaan2023self}, Reflexion~\citep{shinn2023reflexion}, CRITIC~\citep{gou2023critic,peng2025criticlean,zhang2025codecriticbenchholisticcodecritique}) iterate but cannot see, relying on pixel-blind unit tests or linters.

To bridge this gap, we introduce \textbf{ReLook}, a vision-grounded agentic reinforcement learning framework that completes the generate--diagnose--refine loop. The agent actively invokes an MLLM as a tool to "see" rendered outputs and obtain rich textual suggestions during inference, enabling true iterative refinement. Training uses a comprehensive reward system: a powerful MLLM (e.g., Qwen2.5-VL~\citep{wang2024qwen2}) supplies the perceptual signal text-only methods lack, and a rendering-integrity rule assigns zero reward when required screenshots are invalid to deter reward hacking.

However, we identify a critical challenge: behavioral collapse, where despite high-quality feedback, a subsequent revision can be worse. We adopt a Forced Optimization strategy that accepts only strictly improving steps, ensuring high-quality, monotonically improving trajectories. For low-latency inference, the external critic can be discarded; the model performs a lightweight self-edit cycle—render, self-edit, and converge quickly to a human-aligned result.

\paragraph{Evaluator validity and choice.}
Our offline evaluation strictly follows the \textsc{ArtifactsBench} protocol\citep{zhang2025artifactsbench}. \textsc{ArtifactsBench} establishes evaluator validity through: (i) controlled human studies demonstrating over 90\% agreement between MLLM judges (Gemini-2.5-Pro, Qwen2.5-VL-72B) and human experts, and (ii) strong ranking correlation with WebDev Arena\citep{lmsys2024leaderboard}, a large-scale crowdsourced platform. Since our test sets are strict subsets of \textsc{ArtifactsBench}'s evaluated tasks, we directly inherit this established human-alignment evidence. To further mitigate on-policy judge overfitting, we decouple the training-time critic (Qwen2.5-VL-72B-Instruct) from the offline evaluator (Gemini-2.5-Pro). We do not conduct additional human studies; validity rests on \textsc{ArtifactsBench}'s rigorous validation. See Appendix for detailed protocol adherence and cross-judge analysis.

Our contributions are as follows: 
\begin{itemize}
\item \textbf{Robust Reward System.}
    We employ an MLLM as the reward model to provide the rich, pixel-level training signal that text-only methods cannot capture. This is critically supplemented by a zero-reward rule for answers without screenshots, which is designed to prevent reward hacking by forcing the agent to produce renderable code.
\item \textbf{Agent reinforcement learning Framework.}
    We empower the agent to perform a generate–diagnose–refine loop by actively invoking an MLLM as a diagnostic tool. The agent can ``see'' its rendered output and receive rich, actionable feedback for iterative improvement. To ensure this powerful loop is productive and stable, we introduce a Forced Optimization strategy that addresses the challenge of behavioral collapse by guaranteeing the construction of high-quality, monotonically improving rollout trajectories.
\item \textbf{Broad Applicability.}  
    We perform extensive experiments on three widely-used benchmark datasets, and demonstrate that ReLook significantly outperforms the baselines. Moreover, we show the compatibility of ReLook by integrating it with different LLMs.
\end{itemize}

\section{Method}
\label{sec:method}

% \FloatBarrier
\begin{figure*}[htbp]
\centering
\includegraphics[width=0.95\textwidth]{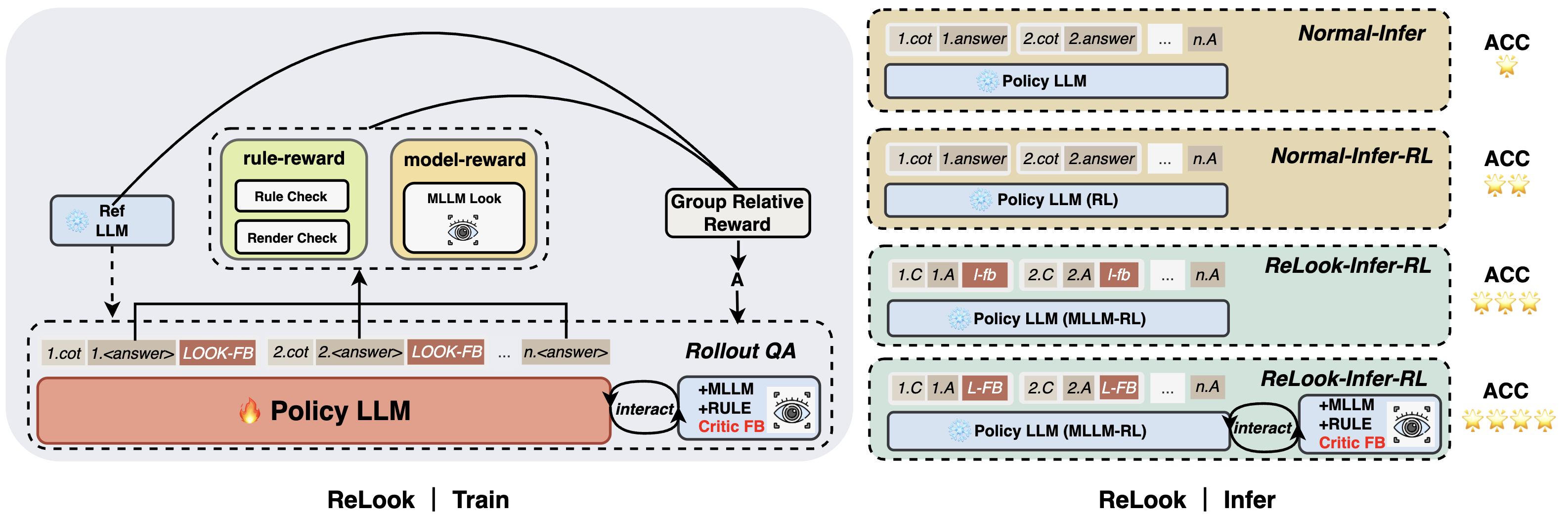}
\caption{Overview of \textbf{ReLook}. Left: training closes a generate–diagnose–refine cycle: policy LLM generates code, pages rendered to temporal screenshots, and a vision-aware critic (MLLM) provides scores and feedback. Rewards combine visual scoring and format constraints; the policy is optimized with GRPO. Right: at inference the model runs a lightweight Re-Look cycle — external critic may be omitted for latency or used for higher accuracy.}

\label{fig:relook_framework}
\end{figure*}

\subsection{Problem Formulation}
The task of front-end code generation is to produce a code snippet $c$, consisting of SVG, HTML, CSS, and JavaScript, that correctly implements a user's intent specified in a natural language question $q$. This code is typically preceded by a textual chain-of-thought, $t$, which outlines the generation plan. As established in the introduction, the correctness of $c$ is not determined by its syntax alone, but by its rendered appearance and behavior. Given the absence of traditional unit tests for front-end code, we propose using an MLLM as the reward model to assess the perceptual quality of the output. Simultaneously, to enable the model to 'see' its rendered results and make improvements, we design an agentic reinforcement learning framework that empowers the agent to invoke the MLLM as a tool for obtaining vision-grounded suggestions for improvement. By doing so, we aim to enhance the front-end code generation capabilities of current LLMs.

\subsection{Overall Framework}
The overall framework of the proposed ReLook is shown in Figure \ref{fig:relook_framework}. At its core, our goal is to empower the agent to 'see' its rendered output and iteratively improve upon it during inference. We achieve this by designing an agentic reinforcement learning framework that establishes a generate--diagnose--refine loop, where the agent learns to invoke an MLLM as a diagnostic tool. To adjudicate the quality of each code revision, we institute a comprehensive reward system centered on a powerful MLLM. Crucially, to prevent the phenomenon of behavioral collapse, where optimizations paradoxically lead to inferior results, we introduce a Forced Optimization strategy. This strategy refines rollouts to construct higher-quality trajectories, thereby instilling a behavioral logic of monotonic improvement in the agent. Ultimately, this robust training allows the agent to internalize its reflective capabilities, enabling the MLLM critic to be discarded during inference to dramatically accelerate the process.

% \subsection{Problem Formulation}
% Given a natural-language query $q$, the model outputs text $t$ and front-end code $c$ (HTML/CSS/JS/SVG). Correctness depends on rendered appearance and behavior rather than syntax. We therefore use an MLLM judge to provide perceptual reward and design an agentic RL framework that allows vision-grounded feedback to refine $c$.

% \subsection{Overall Framework}
% The overall framework of the proposed ReLook is shown in Figure \ref{fig:relook_framework}. ReLook runs a generate--diagnose--refine loop: render to temporal screenshots, score with an MLLM, incorporate feedback, and continue under a \emph{Forced Optimization} rule that accepts only strictly improving steps. Reflection is internalized so the external critic can be dropped at test time.

\subsection{Iterative Reflection Mechanism}
For each query $q$, the policy emits $t$ and $c$. Upon \texttt{<get\_feedback>}, we execute $c$ in a sandbox, capture screenshots, and query the MLLM for feedback $m$ (wrapped in \texttt{<mllm\_feedback>}). We feed $\{q,t,c,m\}$ into the next round and stop when feedback is not requested or a round cap is reached.

The final output is represented as:
\begin{equation}
o = [t_1 \oplus c_1 \oplus m_1 \oplus \dots \oplus t_R \oplus c_R]
\end{equation}
where $t_r, c_r, m_r$ denote the $r$-th round's text, code and feedback blocks, and $R$ is the total number of reflection rounds. The prompt template is provided in the appendix.

% \subsection{ReLook Overview}

% ReLook is a vision\textendash aware RL framework closing a generate\,\textendash\,diagnose\,\textendash\,refine loop. Each iteration: (i) generate $c$, (ii) render and capture $\mathcal{S}$, (iii) query an MLLM critic for a visual score and feedback $m$, (iv) condition the next step on $(q,o,\mathcal{S},m)$. To prevent oscillation and degradation, we accept only steps that strictly improve the best\textendash so\textendash far score.

\subsection{Reinforcement Learning Framework}

To optimize this framework, we employ Group Relative Policy Optimization (GRPO) as our training algorithm, which is based on a token-level policy gradient loss and is related to PPO~\citep{schulman2017ppo} while differing from preference-based objectives such as DPO~\citep{rafailov2023dpo}. The objective is defined as follows:

\begin{equation}
\begin{aligned}
\mathcal{J}_{GRPO}(\theta)  &= \mathbb{E}\left[q \sim P(Q), \{o_{i}\}_{i=1}^{G} \sim \pi_{\theta_{\mathrm{combined}}}(O|q)\right] \\
&= \frac{1}{\sum_{i=1}^{G}|o_i|}\sum_{i=1}^{G}\sum_{t=1}^{|o_i|} \Bigg\{ \min \left[ \frac{\pi_\theta(o_{i,t}|q_{i,t})}{\pi_{\theta_{old}}(o_{i,t}|q_{i,t})} \hat{A}_{i,t}, \right. \\
&  \quad \left. \mathrm{clip}\left( \frac{\pi_\theta(o_{i,t}|q_{i,t})}{\pi_{\theta_{old}}(o_{i,t}|q_{i,t})}, 1-\varepsilon, 1+\varepsilon \right) \hat{A}_{i,t} \right] - \beta \, \mathrm{D}_{KL}\left[\pi_\theta || \pi_{ref} \right] \Bigg\}
\end{aligned}
\label{eq:grpo}
\end{equation}
\noindent Only tokens in $t,c$ contribute non-zero advantages; critic tokens $m$ are masked ($\hat{A}_{i,t}{=}0$ on $m$).

\paragraph{Advantage estimation and credit assignment.}
We sample $G$ trajectories per query and compute returns from $R_{\text{ReLook}}(o_i)$ (Eq.~\ref{eq:relook_reward}). Using the group mean $b$ as baseline, the advantage is $\tilde{A}_i = R_{\text{ReLook}}(o_i) - b$, broadcast to policy tokens (text $t$, code $c$) while masking critic tokens $m$ ($\hat{A}_{i,t}{=}0$ on $m$). Advantages are standardized and clipped to $[-2,2]$. We regularize toward $\pi_{ref}$ with KL weight $\beta$. The combined policy $\pi_{\theta_{\mathrm{combined}}} $ is defined as: $[\pi_{\theta_{\mathrm{old}}}\,\text{for } t_r, c_r]\oplus[\pi_{\mathrm{MLLM}}\,\text{for } m_r]$, where $\pi_{\mathrm{MLLM}}$ is a frozen MLLM critic (Qwen2.5-VL-72B-Instruct).  Trajectories mix policy tokens ($t_n, c_n$) and critic tokens ($m_n$). During training, only $\pi_{\theta}$ is updated. The hyperparameters $\varepsilon$ and $\beta$ control clipping range and KL regularization strength.
% Trajectories mix policy tokens ($t_n, c_n$) and critic tokens ($m_n$); only $\pi_{\theta}$ is updated. 
% We use Qwen2.5\textendash VL\textendash 72B\textendash Instruct as the critic. Here, $\varepsilon$ controls clipping and $\beta$ weights the KL penalty. 

\paragraph{Token masking and lightweight feedback distillation.}
We optimize GRPO over policy tokens ($t$, $c$) and mask critic tokens $m$ by zeroing advantages. To enable critic-free inference, we optionally distill $m$-tokens with lightweight loss \(\mathcal{L}_{\text{distill}}^{m}\) toward frozen MLLM outputs (see Appendix \S\ref{sec:distill} for details). The total objective is
\(\;\mathcal{L} = -\mathcal{J}_{\text{GRPO}}^{t,c} \, + \, \gamma\, \mathcal{L}_{\text{distill}}^{m}\) (default $\gamma{=}0.1$, sweep $\gamma\in[0.05,0.3]$). This lets RL improve visual quality while distillation transfers feedback style for test-time self-reflection.

\subsection{Reward Design}

Conventional RL signals for code, such as unit tests or linters, operate purely in the text domain and are blind to visual defects. To address this, our reward is derived from a powerful Multimodal LLM (Qwen2.5-VL-72B-Instruct) that scores rendered pages based on the user prompt, the generated code, and a series of temporal screenshots. Capturing screenshots at multiple time points allows the critic to assess dynamic behavior. Within each reflection round we capture three time points $\{S_1,S_2,S_3\}$ (e.g., post-load, +1s, +2s) and jointly evaluate them in a single scoring call to obtain one round-level score. To properly credit incremental progress across rounds, we then average the round-level scores within a trajectory.

A critical component of this design is a safeguard against reward hacking, where malformed but syntactically plausible code might be over-rewarded. We enforce a strict renderability constraint: if any required screenshot is invalid (e.g., due to a render failure or timeout), the reward is zero. While this eliminates degenerate reward channels, it can lead to sparse rewards early in training. To mitigate this, we engineer all prompts to include a visual-output constraint that explicitly instructs the agent to write executable HTML/CSS/JavaScript/SVG code suitable for browser rendering (see Appendix for full prompt template). This simple but effective technique increases the initial Valid Render Rate (from ${\sim}40\%$ at the start of training to ${\sim}80\%$ upon convergence) and better aligns the generation task with our vision-grounded reward. This reward function is formally defined as:
\begin{equation}
R_{\text{MLLM}}(o) = \begin{cases} \text{VisualScore}(o) & \text{if screenshot valid} \\ 0 & \text{otherwise} \end{cases}
\label{eq:r_mllm}
\end{equation}
To discourage repetition, we apply a linear length penalty from $L_{start}$ to $L_{end}$ (12k and 14k tokens):
\begin{equation}
R_{\text{len}}(o) =
\begin{cases}
1 & \text{if } \text{len}(o) < L_{\text{start}} \\
\frac{L_{\text{end}} - \text{len}(o)}{L_{\text{end}} - L_{\text{start}}} & \text{if } L_{\text{start}} \le \text{len}(o) \le L_{\text{end}} \\
0 & \text{if } \text{len}(o) > L_{\text{end}}
\end{cases}
\end{equation}
The final training reward is:
\begin{equation}
R_{\text{ReLook}}(o) = R_{\text{MLLM}}(o) \cdot R_{\text{len}}(o)
\label{eq:relook_reward}
\end{equation}

\subsection{Forced Optimization}
While the MLLM critic provides rich, detailed feedback, we observe a critical instability we term behavioral collapse: despite high-quality suggestions, a subsequent revision may score lower than the previous one, degrading trajectories.

We therefore adopt a \textbf{Forced Optimization} mechanism. We initially explored a \emph{negative-reward penalty} for regressions (rejecting worse-than-previous outcomes by penalizing returns), but found it incentivized the agent to \emph{reduce} reflection frequency to avoid penalties, i.e., reward hacking. Hence, we discard the penalty design and enforce a \emph{strict acceptance rule}: a refinement step (new $t_{r+1}$, $c_{r+1}$) is accepted \emph{only if} its reward strictly exceeds the best-so-far in the trajectory (no $\varepsilon$ margin or re-scoring). Non-improving steps are rejected and a new attempt is sampled, with a maximum of 10 resampling attempts per reflection round. If the limit is reached without improvement, we terminate further reflection for that trajectory and use the best-so-far result. This guarantees monotonically improving accepted trajectories and stabilizes learning without suppressing useful reflections.

\subsection{Efficient Inference}
During training we retain the MLLM feedback loop for self-correction. At inference, we drop the external MLLM and run a lightweight, critic-free self-edit (at most three rounds), with screenshots and MLLM calls disabled. This preserves most gains while substantially reducing latency, following a train-slow, run-fast paradigm. See Appendix for pseudocode.

\section{Experiment}
\label{sec:experiment}

\subsection{Experimental Settings}

\paragraph{Training Data Curation.}
We curate a 3{,}000-task corpus of front-end-only tasks, normalize descriptions, and remove near-duplicates via lexical/DOM/code similarity. Prompts are audited to remove hints and sanitized; the final data are split into train/val stratified by UI archetypes.

\paragraph{Train--Test De-duplication Protocol.}
To prevent leakage to ArtifactsBench, FullStack-Bench-Html and Web-Bench, we run instance-level, multi-view de-duplication before training:
\begin{itemize}
\item \textbf{Lexical}: TF--IDF over char 3-grams; cosine $>0.85$.
\item \textbf{DOM}: tag-bigram Jaccard $>0.90$; fallback tree-edit distance for edge cases.
\item \textbf{Code}: token-set Jaccard $>0.90$ after stripping comments/whitespace and minifying.
\end{itemize}
If any criterion triggers, the instance is removed. Borderline cases (e.g., lexical in $[0.80,0.85]$ plus structural overlap) are manually reviewed. The same procedure purges intra-train near-duplicates.

\paragraph{Dataset.} We evaluate the performance of our method on three widely used datasets: ArtifactsBench~\citep{zhang2025artifactsbench}, FullStack-Bench-Html~\citep{Cheng2024FullStackBE} and Web-Bench~\citep{xu2025webbench}. ArtifactsBench contains 1,825 tasks focused on generating dynamic and interactive visual outputs, such as SVG visualizations and mini-games. Its evaluation protocol uses a Multimodal LLM to score the visual fidelity and interactive integrity of the rendered code.
ArtifactsBench utilizes Gemini-2.5-Pro as an expert judge to evaluate model outputs across its 1,825 tasks, demonstrating over 90\% agreement with human evaluators. To manage evaluation costs, we conduct our experiments on six of its sub-datasets. We use the shorthand A-* for ArtifactsBench subsets: A-Lite (300 randomly sampled cases), A-Easy (305 simple frontend cases), A-Game (all 413 game-related cases), A-SVG (all 123 SVG-focused cases), A-Web (all 447 web-specific cases), and A-Si (all 75 simulation-oriented cases).
FullStack-Bench-Html provides a collection of front-end programming tasks where functional correctness is programmatically validated by passing a suite of predefined unit tests.
Web-Bench simulates realistic web development workflows through 50 complex projects of 20 sequential tasks each. Following its official protocol, performance is measured by passing end-to-end test cases that validate the final project's functionality, and we report the pass@2 rate.

\paragraph{Sandboxed Rendering Environment.}
We execute model-produced code within a headless Chromium-based renderer that is both deterministic and secure. This sandboxed environment operates at the OS level, with filesystem and network access disabled. Safety is further enhanced by blocking dangerous APIs (e.g., \texttt{window.open}, \texttt{alert/confirm/prompt}, \texttt{eval}/\texttt{Function}, clipboard access, non-local \texttt{fetch}/\texttt{XMLHttpRequest}/\texttt{WebSocket}), and enforcing a per-sample wall-clock timeout. Requests to non-whitelisted origins are intercepted and failed closed. To ensure determinism, all external resources like fonts and images are replaced with local fixtures; timers and animations use deterministic seeds; and we enforce a strict Content Security Policy that disallows inline scripts and remote scripts. Within this controlled environment, our visual capture mechanism is dynamic: we take full-page screenshots that automatically adjust to the content's full dimensions, guaranteeing no elements are missed. To capture dynamic behavior, we record these screenshots at $T{=}3$ distinct time points (e.g., post-load, +1s, +2s), which are then passed to the MLLM critic to compute the reward signal.

\paragraph{Evaluator validity and cross-judge robustness.}
We adopt the evaluation protocol from \textsc{ArtifactsBench}\citep{zhang2025artifactsbench}, which establishes validity through two key mechanisms: (i) a controlled human study showing over 90\% agreement between MLLM evaluators (Gemini-2.5-Pro and Qwen2.5-VL-72B) and human experts across diverse front-end tasks, and (ii) strong ranking correlation (Spearman $\rho > 0.85$) with WebDev Arena\citep{lmsys2024leaderboard}, a large-scale crowdsourced evaluation platform. Critically, our test sets (A-Lite, A-Easy, A-Game, A-SVG, A-Web, A-Si) are strict subsets of \textsc{ArtifactsBench}'s human-validated tasks, allowing us to directly inherit the established human-alignment evidence. Consistent with \textsc{ArtifactsBench}, we decouple the training-time critic (Qwen2.5-VL-72B-Instruct, open-source) from the offline evaluator (Gemini-2.5-Pro, proprietary). Cross-judge consistency and further details are summarized in Appendix \S\ref{sec:external_validity}.

\subsection{Evaluation Protocol}
\label{sec:eval_protocol}
We follow a pixel-grounded evaluation protocol aligned with \textsc{ArtifactsBench}. For vision-based front-end tasks, models produce code that is executed in our sandboxed browser to capture temporal screenshots at three time points (post-load, +1s, +2s). An independent evaluator (Gemini-2.5-Pro) assigns a VisualScore on the $[0,100]$ scale considering: (i) adherence to the textual specification, (ii) layout alignment and spatial fidelity, (iii) typography and color coherence, and (iv) interactive integrity when actions are specified. If no valid screenshot is produced, the score is set to zero. We report means over three runs (three random seeds) with identical decoding parameters.

For FullStack-Bench-Html, we follow the benchmark's official unit-test protocol and report the pass rate under the same inference setup as other methods. For Web-Bench, we follow its official end-to-end evaluation and report pass@2 across projects. Unless otherwise stated, all reported metrics—including Web-Bench pass@2—are averaged over three runs. Decoding hyperparameters (temperature and top-p) are fixed across systems unless otherwise stated; local fixtures are cached to avoid network variance.

\paragraph{Baselines and variants.} 
Our experiments are conducted on two strong instruction-tuned base models: Qwen2.5-7B-Instruct and Llama-3.1-8B-Instruct. On each of these backbones, we compare three distinct approaches: (i) \emph{Base Model} (the frozen instruction-tuned model, serving as a direct baseline), (ii) \emph{Web-RL} (vision-grounded RL using the MLLM reward but without the agentic reflection mechanism), and (iii) \emph{ReLook} (our full framework with agentic MLLM-in-the-loop reflection). All methods use identical inference parameters, prompts, and rendering infrastructure for fair comparison. We also include results for GPT-4o (via OpenAI API) and Qwen2.5-32B-Instruct (local deployment) as reference points representing stronger base models; these are evaluated under the same protocol. We do not compare with prior specialized visual code generation methods (e.g., Design2Code\citep{si2024design2code}, UICoder\citep{wu2024uicoder}) because: (i) they were evaluated on different datasets without established cross-benchmark protocols, and (ii) official implementations are not publicly available for controlled comparison on our benchmarks. Our Web-RL baseline serves as a strong vision-aware RL reference that isolates the contribution of agentic reflection.

\paragraph{Manual validation (GSB).}
To complement automated evaluation, we conduct a double-blind human study on 100 randomly sampled tasks comparing Qwen2.5-7B-ReLook against Qwen2.5-7B-Instruct. Five independent annotators directly run both systems' code in our sandboxed renderer and select one of three labels: G (ReLook better), S (same), B (ReLook worse). Majority voting aggregates per-task labels. Results are summarized as G:S:B $= 50:30:20$, indicating a clear preference for ReLook under human judgment.

\paragraph{Implementation Details}
We implement ReLook with grouped rollouts under GRPO. The model is trained for a maximum of 40 steps with a training batch size of 256. 
The learning rate is set to 1e-6, and we employ a linear warmup (first 5\%) and cosine decay schedule. For the GRPO loss function, we set the group size $G$ for rollouts per query to 8 and the clipping parameter $\varepsilon$ to 0.2. We apply a KL penalty toward a reference policy with a non-zero weight $\beta$ and sweep $\beta\in[0.01,0.05]$ (step 0.01). We also sweep the advantage clipping bound in $\{1,2,3\}$ for robustness.
The group size and the number of sampled trajectories per query are chosen to fit accelerator memory while maintaining adequate exploration; gradient accumulation is used to emulate larger effective batch sizes. Mixed precision (bfloat16/fp16) and activation checkpointing reduce memory footprint. For critic feedback tokens, GRPO is masked out; we optionally add a lightweight distillation loss with weight $\gamma$ on the feedback tokens to imitate the frozen MLLM's feedback style. Unless otherwise noted, we use $\gamma{=}0.1$ and sweep $\gamma\in[0.05,0.3]$ in sensitivity checks. For the length penalty, we set the bounds to $L_{\text{start}} = 12k$ and $L_{\text{end}} = 14k$.

We use 64 GPUs (32 policy, 32 MLLM), ~80 minutes per step, and a curated corpus. Decoding is identical across systems (temp 1.0, top-p 0.7); results average three seeds. Prompts and sandbox are shared. We select checkpoints by mean VisualScore. We focus on 7B/8B backbones; larger-scale RL is future work.

\subsection{Main Results}

\begin{table*}[t]  % 使用table*环境跨列显示
\centering  % 内容居中
\resizebox{1.0\textwidth}{!}{  % 可调整宽度比例
\begin{tabular}{lcccccccc}
\hline
Model & A-Lite & A-Easy & A-Game & A-SVG & A-Web & A-Si & FullStack-Bench-Html & Web-Bench \\ \hline
Qwen2.5-32B-Instruct & 25.73 & 33.52 & 24.36 & 26.36 & 27.34 & 25.30 & 70.00 & 10.90 \\
GPT-4o & 33.25 & 34.23 & 33.04 & 33.74 & 34.31 & 31.44 & 71.25 & 23.80 \\ \hline
Llama-3.1-8B-Instruct & 21.04 & 27.11 & 19.25 & 20.33 & 21.91 & 19.18 & 57.50 & 2.50 \\
Llama-3.1-8B-Instruct-Web-RL & 21.67 & 29.80 & 20.61 & 21.64 & 23.15 & 20.44 & 61.75 & 2.75 \\
Llama-3.1-8B-Instruct-ReLook-w/o-MLLM & 22.32 & 30.52 & 21.18 & 22.41 & 24.03 & 22.97 & 63.75 & 2.90 \\
Llama-3.1-8B-Instruct-ReLook & \textbf{23.08} & \textbf{31.86} & \textbf{22.04} & \textbf{22.74} & \textbf{25.42} & \textbf{24.52} & \textbf{63.75} & \textbf{2.90} \\ \hline
Qwen2.5-7B-Instruct & 21.59 & 30.70 & 20.62 & 17.70 & 25.69 & 18.61 & 65.00 & 3.00 \\
Qwen2.5-7B-Instruct-Web-RL & 24.89 & 32.64 & 22.14 & 18.87 & 26.73 & 18.82 & 63.25 & 3.48 \\
Qwen2.5-7B-Instruct-ReLook-w/o-MLLM & 25.44 & 33.29 & 23.05 & 18.92 & 27.11 & 22.15 & 67.50 & 4.20 \\
Qwen2.5-7B-Instruct-ReLook & \textbf{27.88} & \textbf{34.12} & \textbf{26.72} & \textbf{20.92} & \textbf{28.31} & \textbf{26.36} & \textbf{67.50} & \textbf{4.20} \\ \hline

\end{tabular}}
\caption{Main results on ArtifactsBench subsets (A-Lite/Easy/Game/SVG/Web/Si), FullStack-Bench-Html, and Web-Bench (pass@2). VisualScores follow \textsc{ArtifactsBench} $[0,100]$ scale. ReLook uses up to 3 reflection rounds; ReLook-w/o-MLLM relies on internalized self-reflection without external critic. For unit-test benchmarks (FullStack, Web-Bench), both variants use identical critic-free inference, yielding same scores. Bold: best per backbone. Means over 3 seeds (temp=1.0, top-p=0.7).}
\label{tab:main_results}
\end{table*}

\begin{figure}[htbp]
    \centering
    \begin{minipage}[t]{0.545\textwidth}
        \centering
        \includegraphics[width=\columnwidth]{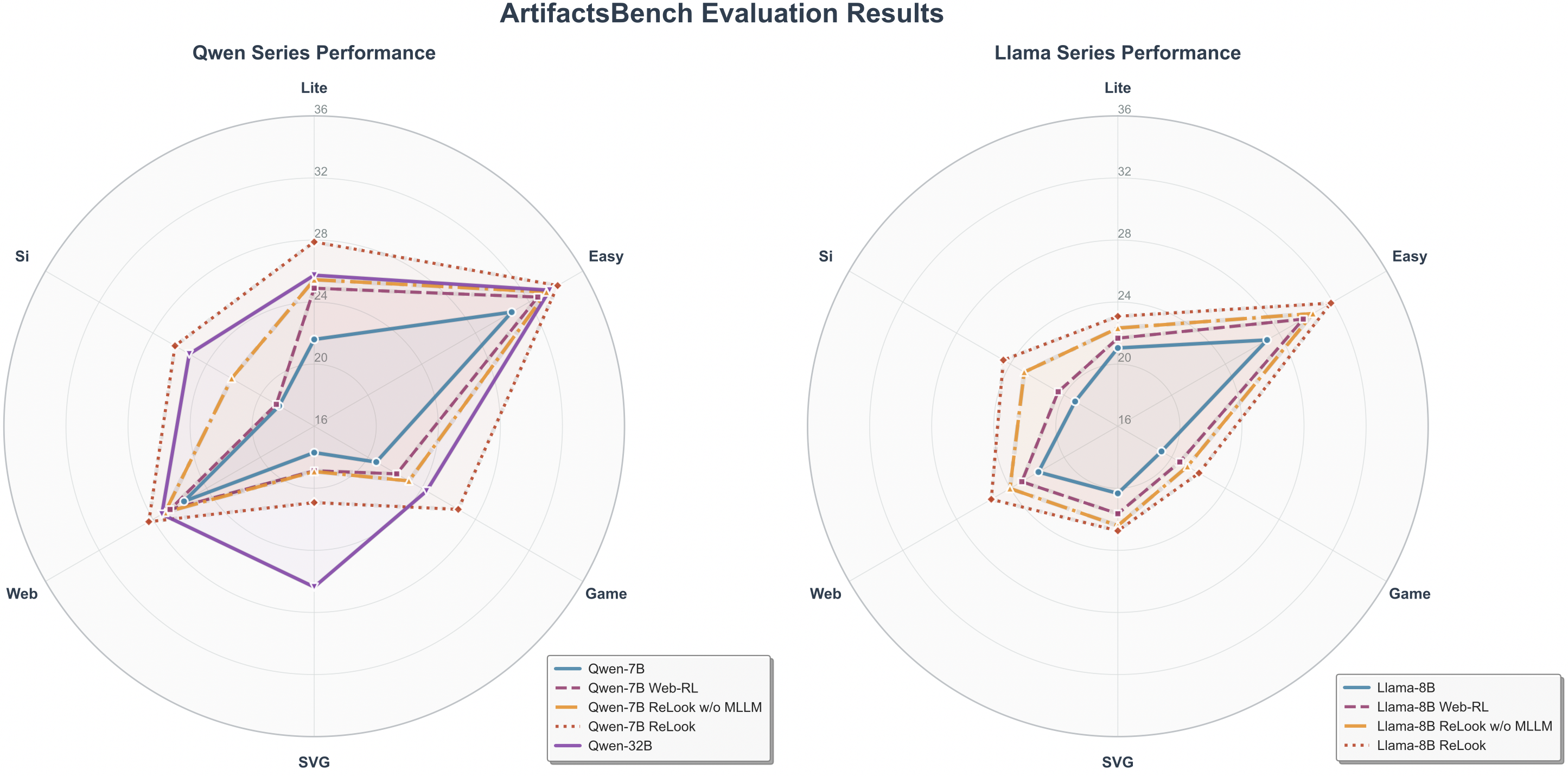}
        \caption{Radar plot showing ReLook's consistent improvements across all ArtifactsBench subsets for both Qwen2.5-7B and Llama-3.1-8B backbones (averaged over 3 seeds).}
        \label{fig:radar}
    \end{minipage}\hfill
    \begin{minipage}[t]{0.425\textwidth}
        \centering
        \includegraphics[width=0.65\columnwidth]{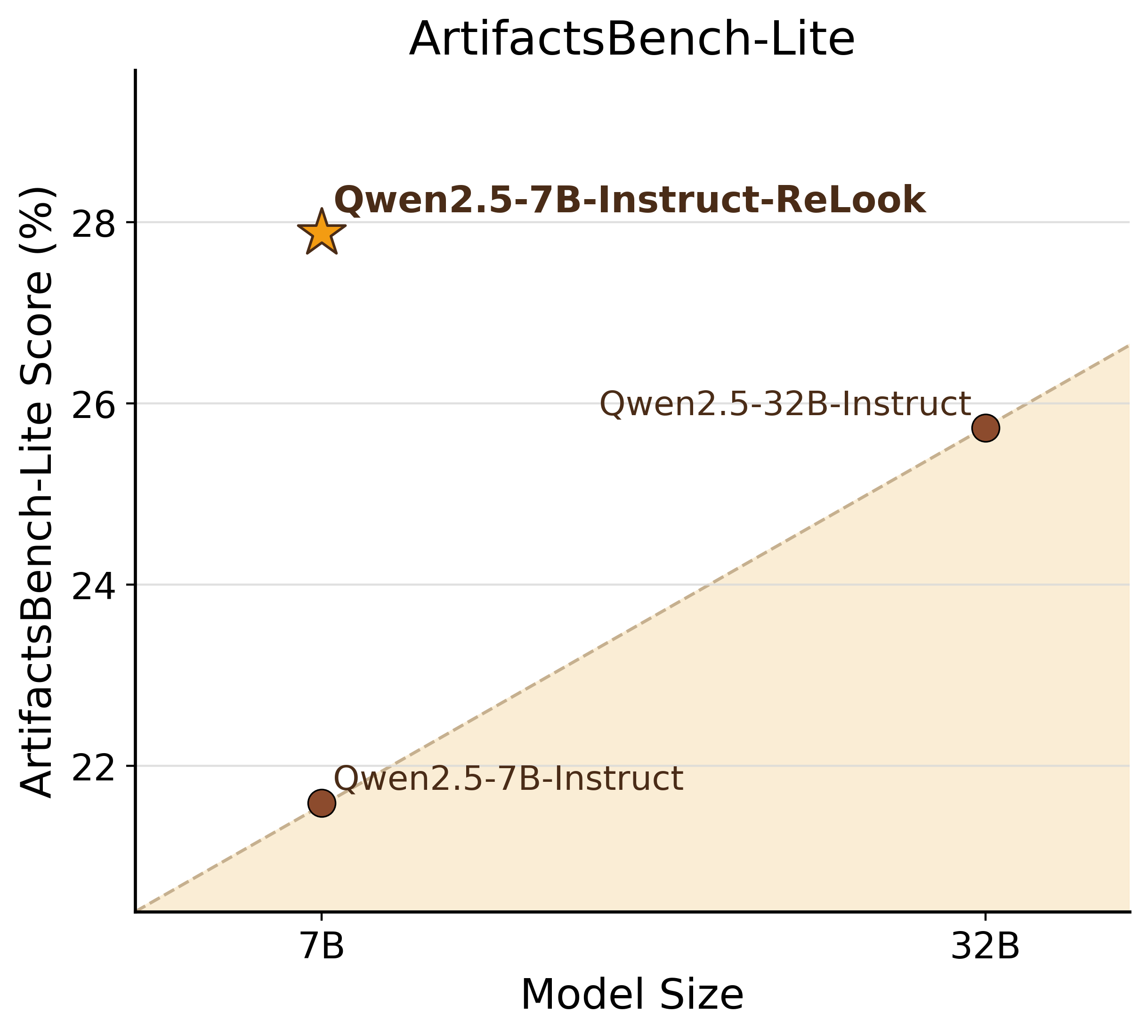}
        \caption{Performance on ArtifactsBench-Lite showing consistent ordering: ReLook $>$ Web-RL $>$ Base Model. Results averaged over 3 seeds.}
        \label{fig:artifactsbench_lite}
    \end{minipage}
\end{figure}

Table \ref{tab:main_results} and Figure \ref{fig:radar} present the main results. ReLook achieves substantial improvements over both base models and Web-RL (vision-grounded RL without agentic reflection). Notably, ReLook-w/o-MLLM—which relies solely on internalized reflection without external critic calls—still outperforms Web-RL, demonstrating successful internalization of the refinement mechanism with minimal inference overhead.

Beyond absolute scores, we consistently observe the strictly monotone ordering predicted by our design: \( \text{ReLook} > \text{Web\text{-}RL} > \text{Base Model}\). The gap between Web-RL and ReLook underscores the importance of a \emph{vision-aware} training signal coupled with the generate--diagnose--refine loop. Qualitative examples are shown in Appendix Figure~\ref{fig:qualitative_analysis}.

Figure \ref{fig:artifactsbench_lite} summarizes performance on the ArtifactsBench-Lite subset. It exhibits the strictly monotone ordering \(\text{ReLook} > \text{Web\text{-}RL} > \text{Base Model}\) and highlights consistent gains for both Qwen2.5-7B and Llama-3.1-8B backbones, echoing the trends in Table \ref{tab:main_results}. This compact subset mirrors the broader benchmark and provides an intuitive visualization of the average improvements delivered by ReLook.

\begin{figure}[htbp]
    \centering
    \begin{minipage}[t]{0.46\textwidth}
        \centering
        \includegraphics[width=0.7\linewidth]{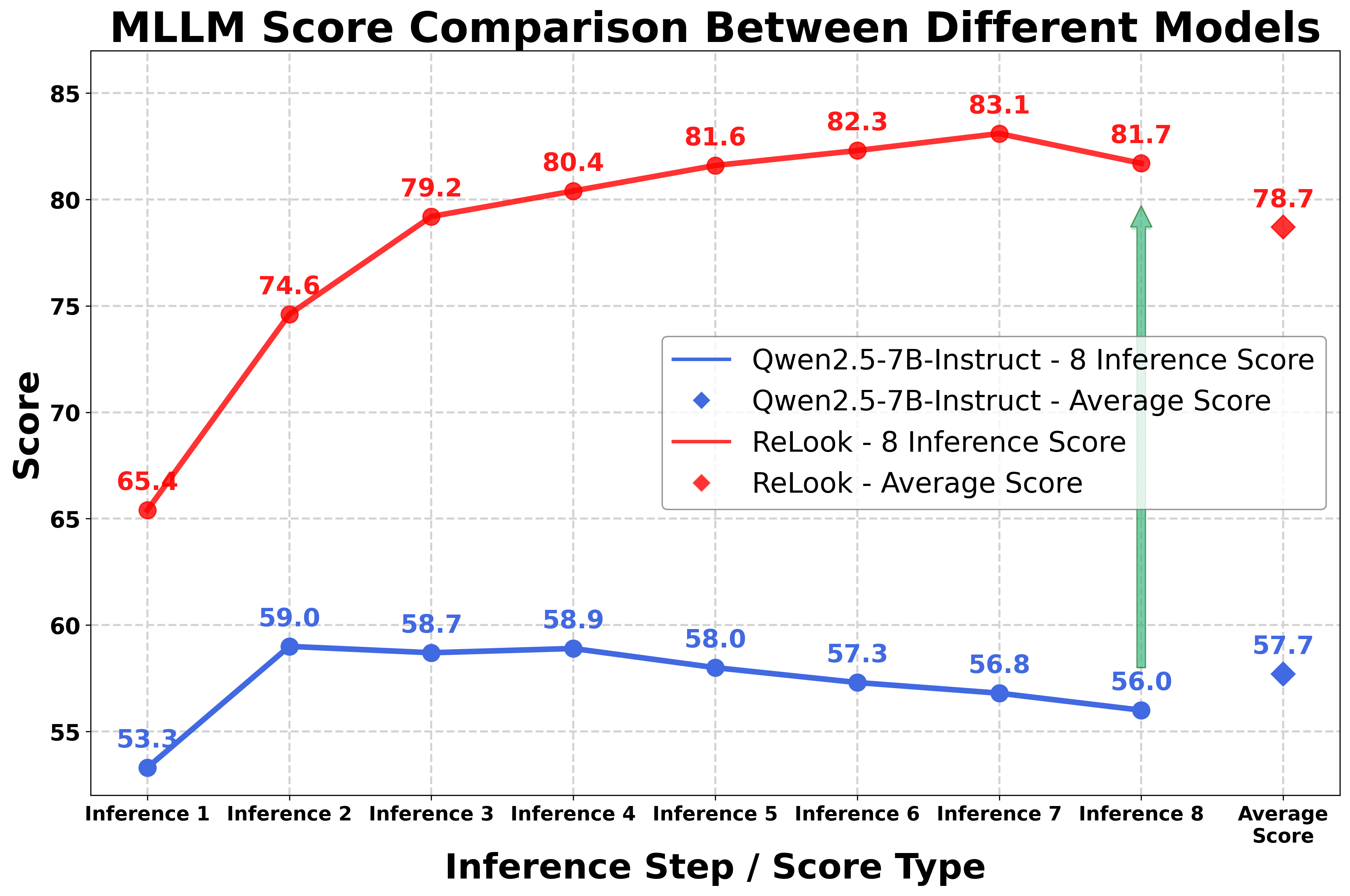}
        \caption{Behavioral collapse mitigation. Base model (Qwen2.5-7B-Instruct) degrades after initial attempts despite MLLM feedback, while ReLook exhibits monotonic improvement across eight forced reflection rounds on ArtifactsBench-Lite. Scores from training-time judge (Qwen2.5-VL-72B).}
        \label{fig:infer3}
    \end{minipage}\hfill
    \begin{minipage}[t]{0.50\textwidth}
        \centering
        \includegraphics[width=\linewidth]{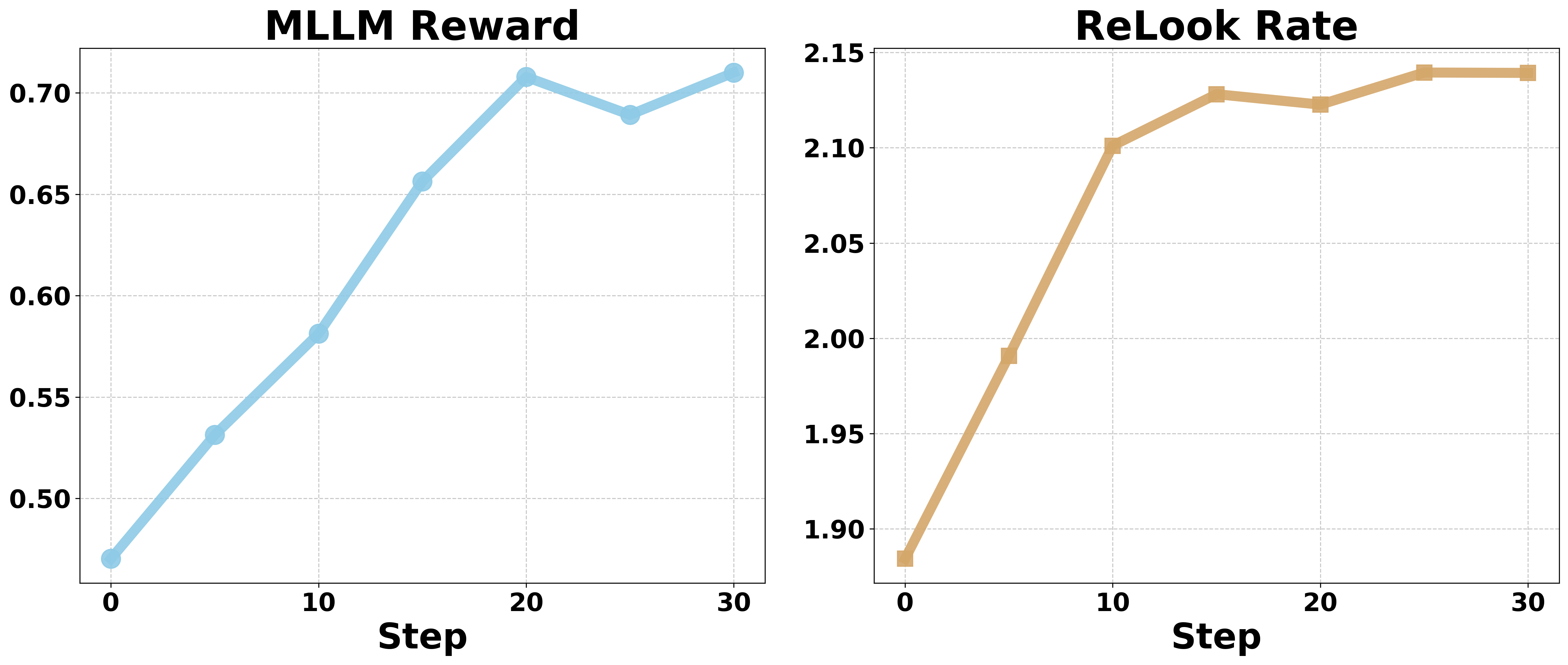}
        \caption{Intermediate Results of RL Training. The figure shows the average reward score on the validation set and the number of optimization steps during inference for our training of Relook using Qwen2.5-Instruct-7B as the base model.}
        \label{fig:mid_result}
    \end{minipage}
\end{figure}

Figure \ref{fig:infer3} evidences behavioral collapse in the base model after 2--3 rounds despite feedback, while ReLook improves monotonically. We cap reflections at three rounds for efficiency and keep this at inference.

Figure \ref{fig:mid_result} shows that reflection frequency increases and stabilizes during training, aligning with Forced Optimization's incentive structure. The average converging to ~2 rounds suggests most tasks reach capacity after two refinements.

\subsection{Ablation Study}
We ablate three components: (i) vision-based MLLM reward, (ii) format constraint invalidating non-renderable outputs, and (iii) Forced Optimization. 
Table~\ref{tab:ablation_study} shows each component is critical. Vision reward provides essential perceptual signal (+3.3 points). Format constraint prevents reward hacking (+1.0). Forced Optimization has the largest impact (+2.0), directly mitigating behavioral collapse. All ablations use identical seeds and the external evaluator to avoid bias.

\begin{table}[htbp]  % 使用table*环境跨列显示
\centering  % 内容居中
\resizebox{0.8\columnwidth}{!}{
\begin{tabular}{cccc}
\hline
MLLM Reward & Format Constraint & Forced Optimization &  ArtifactsBench-Lite~$\uparrow$  \\ \hline
 &  &  & 21.59  \\ 
\checkmark &  &  & 24.89  \\ 
\checkmark & \checkmark &  & 25.84  \\
\checkmark & \checkmark & \checkmark & 27.88  \\ \hline

\end{tabular}}
\caption{Ablation Study on ArtifactsBench-Lite. Results are averaged over 3 random seeds.}
\label{tab:ablation_study}
\end{table}

\subsection{Inference speed improvement after removing MLLM}
We use VLLM to deploy the Qwen2.5-7B-Instruct model on four H20 GPUs and the Qwen2.5-VL-72B-Instruct model on eight H20 GPUs. We set the number of parallel threads to 1 and limit the maximum number of reflection steps to 3. We conduct inference on 100 queries and measure the average inference time per query. ReLook takes an average of 123.04 seconds per query, whereas ReLook-w/o-MLLM takes only 18.03 seconds. The results indicate that removing the screenshot and MLLM invocation mechanism substantially improves inference efficiency.

\subsection{Error Analysis and Task-Level Performance}
We analyze ReLook's improvements across different task types to understand where vision-grounded RL is most effective. Visual-centric tasks (e.g., A-SVG, A-Game, A-Web) show the largest gains: ReLook improves by 3.2-6.1 points over base models on these subsets, as the MLLM critic directly addresses layout precision, color fidelity, and animation dynamics—failures invisible to text-only methods. On A-Easy (simpler static pages), gains are modest (2.4-3.4 points), as base models already achieve reasonable outputs. The most challenging scenario remains complex, multi-file project tasks like Web-Bench, where ReLook shows improvement (40\% relative gain for Qwen2.5-7B: 3.00 $\rightarrow$ 4.20 pass@2) but absolute performance stays low, indicating that long-horizon reasoning and cross-file dependencies require further architectural advances beyond single-artifact refinement.

\subsection{Qualitative Analysis}
Appendix Figure \ref{fig:qualitative_analysis} compares baseline and ReLook outputs. Non-vision baselines exhibit: (i) layout drift (misaligned components), (ii) interaction breakage (missing event listeners), and (iii) aesthetic inconsistency (clashing colors). ReLook mitigates these through render-aware refinements guided by visual critique.

\section{Related Work}
\label{sec:related_work}

\subsection{Visual Code Generation}
Early vision-to-code systems translate static screenshots to HTML/CSS~\citep{wust2024pix2code} in one shot. Recent methods add structure—Design2Code~\citep{si2024design2code}, Web2Code~\citep{yun2024web2code}, UICoder~\citep{wu2024uicoder}, DesignCoder~\citep{chen2025designcoder}—but mainly optimize static similarity, struggling with dynamics and iterative refinement. ReLook trains with a renderer in the loop, uses vision-grounded rewards from temporal screenshots, and internalizes refinement via RL, aligning with cross-modal supervision~\citep{feng2022beyond}.

\subsection{Feedback-Driven Code Reinforcement Learning}
RL for program synthesis leverages unit-test rewards and large candidate sets~\citep{le2022coderl,li2022competition}; reflective/tool-driven critiques improve correction~\citep{madaan2023self,shinn2023reflexion,gou2023critic,peng2025criticlean}. For front-end, unit tests are pixel-blind; even with structured visual instructions~\citep{yun2024web2code}, pixel signals are missing. LLM critics help surface model errors~\citep{mcaleese2024llm}. We couple the policy with a visual reward from temporal screenshots and stabilize training via Forced Optimization and zero-reward for invalid renders.

\paragraph{Acceptance criteria, best-of-N, and verifier-assisted search.}
A broad line of work improves generation via external selection/search: Codex/AlphaCode sample-and-test~\citep{chen2021evaluating,li2022competition}; self-consistency/tree deliberation aggregate candidates~\citep{wang2022selfconsistency,yao2023treeofthought}; agentic self-refinement uses iterative critique~\citep{madaan2023self,shinn2023reflexion,gou2023critic, huang2025think}. Our \emph{Forced Optimization} differs in both \emph{criterion} and \emph{rule}: a vision-grounded score from temporal screenshots (not unit tests), and acceptance \emph{strictly} requiring monotonic improvement within one trajectory. Unlike best-of-$N$ or offline re-ranking, our rule is \emph{online, in-trajectory}, preventing regressions and reward hacking, yielding stable, visually aligned improvements for front-end code.
\subsection{Multimodal UI Perception and Evaluation}
Recent MLLMs ground web elements and layouts~\citep{openai2023gpt4v,wang2024qwen2}; web-agent/GUI benchmarks show the value of vision-conditioned reasoning~\citep{zhou2023webarena,koh2024visualwebarena,li2025screenspot}. For evaluation, MLLM-as-Judge is common~\citep{ge2023mllm,zheng2023judging}; front-end benchmarks emphasize visual and interactive quality~\citep{xu2025webbench,zhang2025artifactsbench}. Accordingly, we place visual scoring in training to internalize layout/interaction principles, and drop the critic at inference for latency.

\medskip
ReLook unifies these threads via MLLM-based visual rewards within RL, offering a practical path to visually aware, self-improving front-end generation.
\section{Conclusion}
We introduced \textbf{ReLook}, a vision-grounded RL framework that closes a generate--diagnose--refine loop for front-end code. By coupling a multimodal LLM critic with two safeguards—zero-reward for invalid renders and Forced Optimization—ReLook achieves consistent gains over strong baselines (\(\text{ReLook} > \text{Web\text{-}RL} > \text{Base}\)) and enables critic-free inference for substantial speedups. We expect this blueprint—placing a perception-aligned evaluator inside the learning loop—to generalize beyond web UIs to other perceptual programming domains. See Appendix for limitations and future directions.

\clearpage
\bibliography{relook}
\bibliographystyle{relook}

\clearpage
\appendix
\clearpage

\section{Limitations and Future Work}
Despite strong empirical gains, ReLook has several limitations. First, the reliance on a large MLLM critic during training increases compute and monetary cost; although we show critic-free inference, reducing training-time overhead via distillation or lighter judges remains future work. Second, our sandboxed renderer improves determinism but may under-represent real-world variability across devices, locales, and resource conditions, leaving robustness gaps. Third, rewards are mediated by an external evaluator and a length penalty; metric drift and sensitivity to prompt templates could bias optimization. Fourth, the Forced Optimization constraint stabilizes training but might reduce exploration in challenging cases. Fifth, as shown in our error analysis (Section~\ref{sec:experiment}), complex multi-file project tasks (e.g., Web-Bench) remain challenging, indicating that long-horizon reasoning beyond single-artifact refinement requires further architectural innovation. Finally, our experiments focus on 7B/8B-scale models and front-end code; scaling to larger models and broader software stacks requires additional investigation. We follow and inherit \textsc{ArtifactsBench}'s cross-judge and human-alignment evidence rather than re-running additional human studies in this work. To facilitate reproducibility and future research, we plan to release our code upon publication.

\section{External validity and cross-judge analysis}
\label{sec:external_validity}
We align our evaluator setup with \textsc{ArtifactsBench}\citep{zhang2025artifactsbench}, which establishes validity through rigorous empirical validation:

\paragraph{Human-MLLM Agreement.} \textsc{ArtifactsBench} conducted a controlled human study with expert web developers evaluating a stratified sample of 200 tasks. Results showed over 90\% agreement (Cohen's $\kappa > 0.85$) between human judgments and MLLM evaluators (Gemini-2.5-Pro and Qwen2.5-VL-72B) on visual fidelity, layout correctness, and interactive integrity. The study used a double-blind protocol with three independent human raters per task.

\paragraph{Crowdsourced Validation.} Beyond controlled studies, \textsc{ArtifactsBench} validated MLLM judges against WebDev Arena\citep{lmsys2024leaderboard}, a large-scale platform with over 10,000 pairwise comparisons from web developers. The MLLM rankings showed strong correlation (Spearman $\rho = 0.87$ for Gemini-2.5-Pro, $\rho = 0.83$ for Qwen2.5-VL-72B) with crowd preferences.

\paragraph{Our Protocol Adherence.} Since our test sets (A-Lite, A-Easy, A-Game, A-SVG, A-Web, A-Si) are strict subsets of \textsc{ArtifactsBench}'s 1,825 human-validated tasks, we directly inherit this established validity. We use the official evaluation scripts, judge prompts, and scoring rubric without modification. To mitigate on-policy overfitting, we decouple training-time critic (Qwen2.5-VL-72B-Instruct) from offline evaluator (Gemini-2.5-Pro). We do not re-run human studies; validity is anchored in \textsc{ArtifactsBench}'s reported evidence.

\paragraph{Score Interpretation.} The absolute VisualScore range (20-30 for 7B/8B models) reflects benchmark difficulty: \textsc{ArtifactsBench} tasks span complex interactions, dynamic animations, and pixel-perfect layouts. Reference models (GPT-4o: 33, Qwen2.5-32B: 26) establish that even frontier systems find these tasks challenging. The $[0,100]$ scale provides fine-grained discrimination; we report relative improvements over baselines.

\section{Pseudocode for ReLook}

\begin{algorithm}[htbp]
\caption{ReLook Training Framework}
\label{alg:relook}
\textbf{Require:} Task specification $q$, base policy $\pi_{\theta}$, vision critic MLLM (Qwen2.5-VL-72B), max steps $S$, max reflections $R$, max resampling $K{=}10$.\\
\textbf{Ensure:} Optimized policy $\pi^*_{\theta}$ with internalized visual cognition.

\begin{enumerate}[leftmargin=*,itemsep=2pt]
\item Initialize: $s \gets 0$, $history \gets [q]$.
\item While $s < S$:
  \begin{enumerate}[leftmargin=*,itemsep=1pt]
  \item Initialize: $r \gets 0$, $o \gets \epsilon$, $s_{\text{prev}} \gets -1$.
  \item While ($r < R$ and room for improvement):
    \begin{enumerate}[leftmargin=*,itemsep=1pt]
    \item $k \gets 0$, $accepted \gets \text{False}$.
    \item While ($k < K$ and not $accepted$): \textit{(Forced Optimization: resample up to $K$ times)}
      \begin{enumerate}[leftmargin=*,itemsep=1pt]
      \item \textbf{Generate:} $t_r, c_r \gets \pi_{\theta}(\text{next token} \mid history)$.
      \item Extract code $c_r$ from \texttt{<answer>} block.
      \item Attempt rendering $c_r$ $\rightarrow$ capture screenshots $\{S_1, S_2, S_3\}$.
      \item Generate feedback $m_r \gets \text{MLLM}(q, c_r, \{S_i\})$.
      \item Compute visual score $s_r \gets \text{VisualScore}(q, c_r, \{S_i\})$.
      \item If $s_r > s_{\text{prev}}$ (accept only strictly improving steps): append $t_r, c_r$ to $history$ and $o$; append $m_r$ to $history$ and $o$ within \texttt{<mllm\_feedback>}; set $s_{\text{prev}} \gets s_r$, $accepted \gets \text{True}$.
      \item $k \gets k + 1$.
      \end{enumerate}
    \item If not $accepted$: break. \textit{(Terminate further reflections, use best-so-far)}
    \item $r \gets r + 1$.
    \end{enumerate}
  \item Calculate reward components for the trajectory $o$:
    \[
    R_{\text{MLLM}}(o) = \begin{cases} \text{VisualScore}(o) & \text{if screenshot valid} \\ 0 & \text{otherwise} \end{cases}
    \]
    \[
    R_{\text{len}}(o) =
    \begin{cases}
    1 & \text{if } \text{len}(o) < L_{\text{start}} \\
    \dfrac{L_{\text{end}} - \text{len}(o)}{L_{\text{end}} - L_{\text{start}}} & \text{if } L_{\text{start}} \le \text{len}(o) \le L_{\text{end}} \\
    0 & \text{if } \text{len}(o) > L_{\text{end}}
    \end{cases}
    \]
    \[
    R_{\text{ReLook}}(o) = R_{\text{MLLM}}(o) \cdot R_{\text{len}}(o)
    \]
  \item Perform policy update using GRPO to optimize parameters $\theta$ based on $R_{\text{ReLook}}$.
  \item If convergence criterion is met (no significant improvement), exit loop.
  \item $s \gets s + 1$.
  \end{enumerate}
\end{enumerate}
\end{algorithm}
\section{Evaluator rubric and scoring range}
We use a fixed evaluation rubric for \textsc{VisualScore}. During training, the critic's visual score used for RL is normalized to $[0,1]$ for stability. For offline reporting and tables, we follow \textsc{ArtifactsBench} and report evaluator outputs on a $[0,100]$ scale; all numbers in the main results tables and figures are on this $[0,100]$ scale unless otherwise specified. The rubric jointly considers: (i) specification adherence; (ii) layout alignment and spatial fidelity; (iii) typography and color coherence; and (iv) interactive integrity for tasks specifying actions. For dynamic behavior, each reflection round is evaluated on three temporal screenshots jointly in a single scoring call to obtain one round-level score; trajectory-level reward averages round-level scores. We do not perform multi-judge averaging or smoothing.

\section{The loss function for Forced Optimization}
Faced with the "behavior collapse" problem, we first attempted a negative-reward penalty comparing post-feedback scores to pre-feedback scores. If an optimized answer was worse than its predecessor, a negative reward was applied. We observed that this encouraged the model to reduce reflection frequency to avoid penalties (reward hacking), degrading ReLook into regular RL and harming performance. Therefore, we removed this mechanism and adopted the strict acceptance rule described in Method: only strictly improving steps are accepted into trajectories.

\section{Lightweight distillation loss on feedback tokens}
\label{sec:distill}
Let $M$ denote the index set of critic feedback tokens $m$ within a trajectory, and let the frozen MLLM critic define a teacher distribution $q_t(\cdot)$ at each position $t\in M$. The student policy distribution is $p_{\theta}(\cdot\mid x_{\le t})$. The lightweight distillation loss used to transfer the feedback style for test-time self-reflection is
\begin{align}
\mathcal{L}_{\text{distill}}^{m}
&:= \frac{1}{|M|} \sum_{t\in M} \mathrm{KL}\big( q_t(\cdot) \,\|\, p_{\theta}(\cdot\mid x_{\le t}) \big)\label{eq:distill_kl}
\end{align}
 equivalently, as a teacher-forced cross-entropy
\begin{align}
\mathcal{L}_{\text{distill}}^{m}
&:= -\frac{1}{|M|} \sum_{t\in M} \sum_{v} q_t(v)\, \log p_{\theta}(v\mid x_{\le t})\label{eq:distill_ce}
\end{align}
Only policy-controlled tokens $(t,c)$ contribute non-zero advantages in GRPO; feedback tokens $m$ are masked in the RL objective (advantages set to zero). The total training objective is
\begin{align}
\mathcal{L}
&= -\mathcal{J}_{\text{GRPO}}^{t,c} + \gamma\, \mathcal{L}_{\text{distill}}^{m}, \quad \text{with default } \gamma{=}0.1,\; \gamma\in[0.05,0.3].\label{eq:total_loss}
\end{align}

\section{Reproducibility notes: GRPO and implementation details}
\paragraph{Reference policy and KL.} We regularize toward a fixed reference policy $\pi_{ref}$ (the frozen instruction-tuned backbone before RL). KL is computed token-wise over policy-controlled tokens with weight $\beta$ (see ranges in Experiment); we do not anneal $\beta$ within a step.
\paragraph{Trajectory composition and masking.} Trajectories mix policy tokens ($t, c$) and critic tokens ($m$). During optimization, advantages on $m$ are masked to zero; optional lightweight distillation on $m$ uses a small KL/CE with weight $\gamma$ to the frozen MLLM outputs.
\paragraph{Advantage baseline and clipping.} We use a group-relative baseline (mean return over $G$ trajectories per query); advantages are standardized within-batch and clipped to $[-2,2]$.
\paragraph{Sampling limits and rejection handling.} For Forced Optimization, we cap resampling attempts at 10 per reflection round to avoid infinite loops; non-improving proposals are rejected without re-scoring. When the 10-attempt limit is reached without improvement, we terminate further reflection rounds for that trajectory and use the best-so-far result. This limit balances exploration (allowing sufficient attempts to find improving revisions) with computational efficiency.
\paragraph{Length penalty parameters.} We set $L_{\text{start}} = 12{,}000$ and $L_{\text{end}} = 14{,}000$ tokens based on empirical analysis of typical front-end code lengths in our training corpus, ensuring reasonable generation length while discouraging degenerate repetition.
\paragraph{Valid Render Rate dynamics.} During training, the Valid Render Rate increases from approximately 40\% at initialization to approximately 80\% upon convergence, demonstrating that the model learns to produce syntactically valid and renderable code through the combination of the visual-output constraint in prompts and the zero-reward penalty for invalid renders.
\paragraph{Decoding and seeds.} Unless otherwise noted, decoding uses identical hyperparameters across systems (temperature=1.0, top-p=0.7) with three random seeds; fixtures are cached to avoid network variance.
These notes complement Section~\ref{sec:method} and Section~\ref{sec:experiment} to facilitate faithful reimplementation.

\section{Qualitative Analysis}

\begin{figure*}[htbp]
\centering
\includegraphics[width=0.75\textwidth]{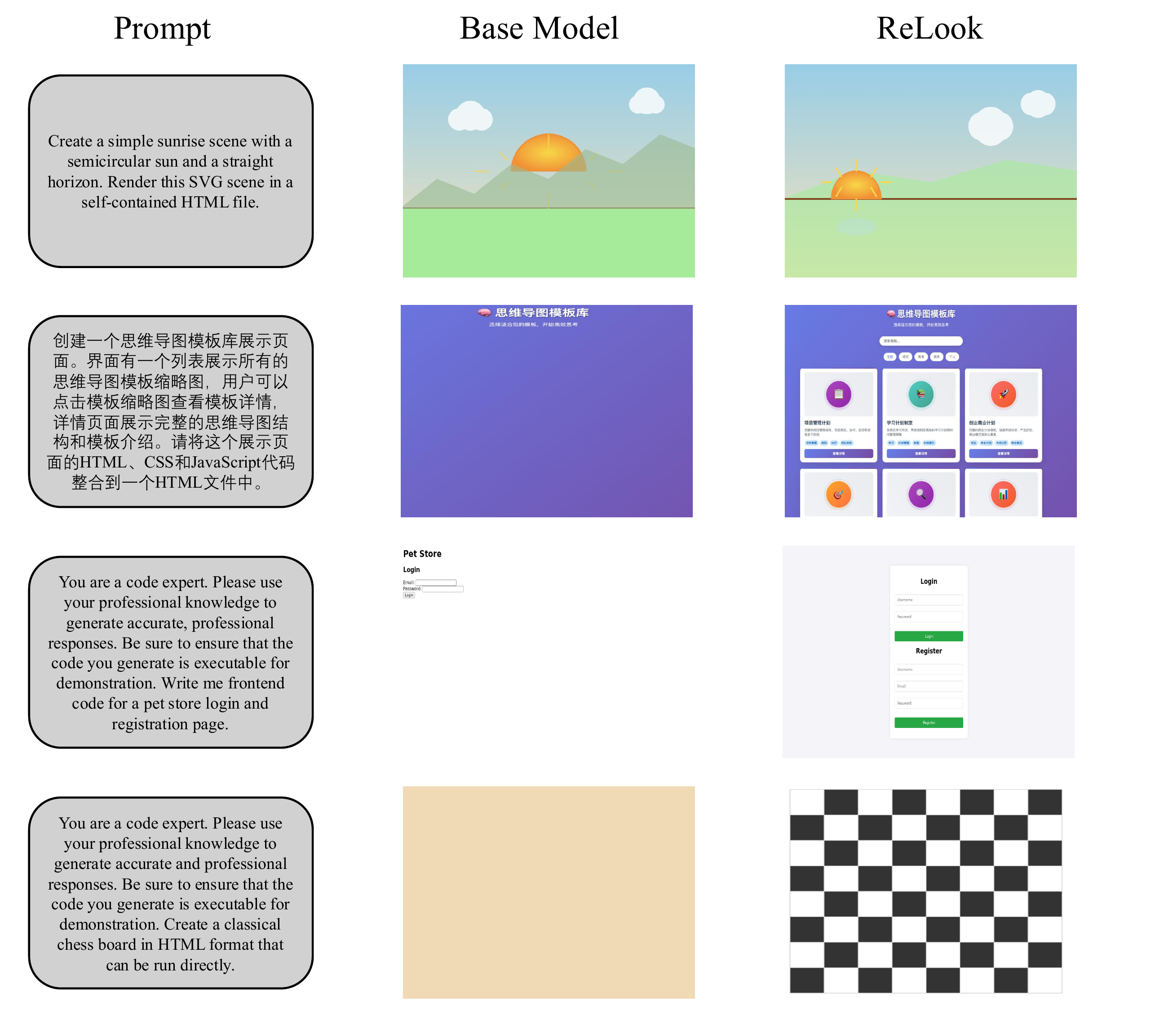}
\caption{Visual Comparison of Frontend Websites Generated by Baseline and ReLook.}
\label{fig:qualitative_analysis}
\end{figure*}

Figure \ref{fig:qualitative_analysis} presents a qualitative comparison between the baseline model and ReLook across several representative front-end generation tasks.

Prompt column lists natural-language instructions of varying complexity, including layout composition, template library rendering, login/registration forms, and a chessboard.

Base Model column shows the outputs of an instruction-tuned baseline. While it can produce code that renders without error, the resulting webpages often suffer from issues such as layout drift, incomplete functionality, and lack of visual coherence (e.g., missing interactivity in the login page, overly simplistic rendering of the chessboard).

ReLook column demonstrates the effect of our vision-grounded reinforcement learning framework. By incorporating visual feedback into training, ReLook produces outputs that are not only executable but also visually faithful and functionally aligned with the prompts. For instance, the template library page is correctly populated with clickable cards, the login/registration form has a clean layout and interactive elements, and the chessboard renders with precise alignment.

Overall, the comparison highlights how ReLook systematically reduces layout drift, strengthens interaction correctness, and achieves higher visual fidelity compared to the baseline.

\section{Template prompt for ReLook rollout}

\begin{figure*}[htbp]
\begin{tcolorbox}
\footnotesize
\begin{CJK}{UTF8}{gbsn}
Solve the following problem step by step. \\
You now have the ability to selectively write executable HTML, CSS, JavaScript, or SVG code to receive feedback from the multimodal large model on the code. \\
The code you provided will be executed, and the feedback (wrapped in '\texttt{<mllm\_feedback>} output\_str \texttt{<mllm\_feedback>}') can be returned to aid your reasoning and help you arrive at the final answer.\\
Unless you believe the current answer is flawless, please output \texttt{<get\_feedback>} after providing the complete answer to receive feedback from the multimodal large model and improve the code based on the feedback.\\
*user question:*\\
\{\$query\}\\
\end{CJK}
\end{tcolorbox}
\caption{Template prompt for ReLook rollout.}
\label{fig:filter_query}
\end{figure*}

\end{document}